%% file: VisappPaper.tex
\documentclass[a4paper,twoside]{article}

\input{Preamble.tex}

\begin{document}

	\title{Procam Calibration from a Single Pose of a Planar Target}
	
	 \author{\authorname{Ghani O. Lawal and Michael Greenspan}
	 \affiliation{Dept. Electrical and Computer Engineering,
	 Ingenuity Labs, Queen's University, Kingston, Ontario, Canada}
	 \email{\{ghani.lawal, michael.greenspan\}@queensu.ca}
	 }
	
	

	
	\abstract{A novel user friendly method is proposed for calibrating a procam system from a single pose of a planar chessboard target.  The user simply needs to orient the chessboard in a single appropriate pose. A sequence of Gray Code patterns are projected onto the chessboard, which allows correspondences between the camera, projector and chessboard to be automatically extracted. These correspondences are fed as input to a nonlinear optimization method that models the projection of the principle points onto the chessboard, and accurately calculates the intrinsic and extrinsic parameters of both the camera and the project, as well as the camera’s distortion coefficients.  The method is  experimentally validated on a real procam system, which is shown to be comparable in accuracy with existing multi-pose  approaches. The impact of the  orientation of the chessboard with respect to the procam imaging places is also explored through extensive simulations. 
	}
	
	\onecolumn \maketitle \normalsize \setcounter{footnote}{0} \vfill
	
	\input{Introduction.tex}
	
	\input{Background.tex}
	
	\input{Method.tex}
	
	\input{Experimentation.tex}

	\input{Conclusion.tex}
	
	
	
	
	
	\section*{Acknowledgements}
	The authors would like to acknowledge and thank Epson Canada, the Natural Sciences and Engineering Research Council of Canada, and the Ontario Centres of Excellence, for their support of this work.
	
	\bibliographystyle{apalike}
	{\small
		\bibliography{References}}
	
	
	
\end{document}

%% file: Preamble.tex
\usepackage{epsfig}
\usepackage{subcaption}
\usepackage{calc}
\usepackage{amssymb}
\usepackage{amstext}
\usepackage{amsmath}
\usepackage{amsthm}
\usepackage{multicol}
\usepackage{pslatex}
\usepackage{apalike}
\usepackage{verbatim}
\usepackage{SCITEPRESS}     

%% file: Introduction.tex
\section{\uppercase{Introduction}}
\label{sec:introduction}

Projector camera (procam) systems are an effective approach for implementing range sensing and interacting with a 3D environment.  For static scenes, the procam system can be used as an inexpensive 3D scanner \cite{paper:GreyLocalHom}.  They can also be used in dynamic settings where the camera captures information about the environment in real time and the projector  displays any visual content that the user provides.  This allows the system to implement Spatial Augmented Reality (SAR) \cite{paper:Fiducials}.  This type of projection-based augmented reality (AR) enables the user to interact with the environment naturally without the requirement to have devices attached to their bodies as in head mounted or handheld AR \cite{paper:DotsRand}. Shader lamps, smart projectors and augmented paintings on non-planar surfaces are a few of the specific applications of SAR achieved through procam systems \cite{book:SAR}.  

Prior to use, the intrinsic and extrinsic parameters of the procam system must be acquired, which  is a process referred to as \emph{calibration}. Projectors are similar to cameras with respect to their system geometry, except that they emit light rather than absorbing it,  which allows many techniques developed for camera calibration to be adapted for calibrating a projector, provided that methods such as structured light are able to obtain camera-projector pixel correspondences.  This work makes use of Gray code patterns to accomplish this task \cite{paper:AllSL}.  Through the use of structured light, the difficulty in establishing camera-projector pixel correspondences is largely alleviated and becomes computationally trivial, unlike passive camera-camera stereovision systems.  However the issue remains that to calibrate a procam system, 3D correspondences need to be established either by moving a planar target to multiple poses, or by making use of a 3D calibration target that possesses specific shape and detail requirements.
\begin{figure}
	\centering
	\includegraphics[width=0.45\textwidth]{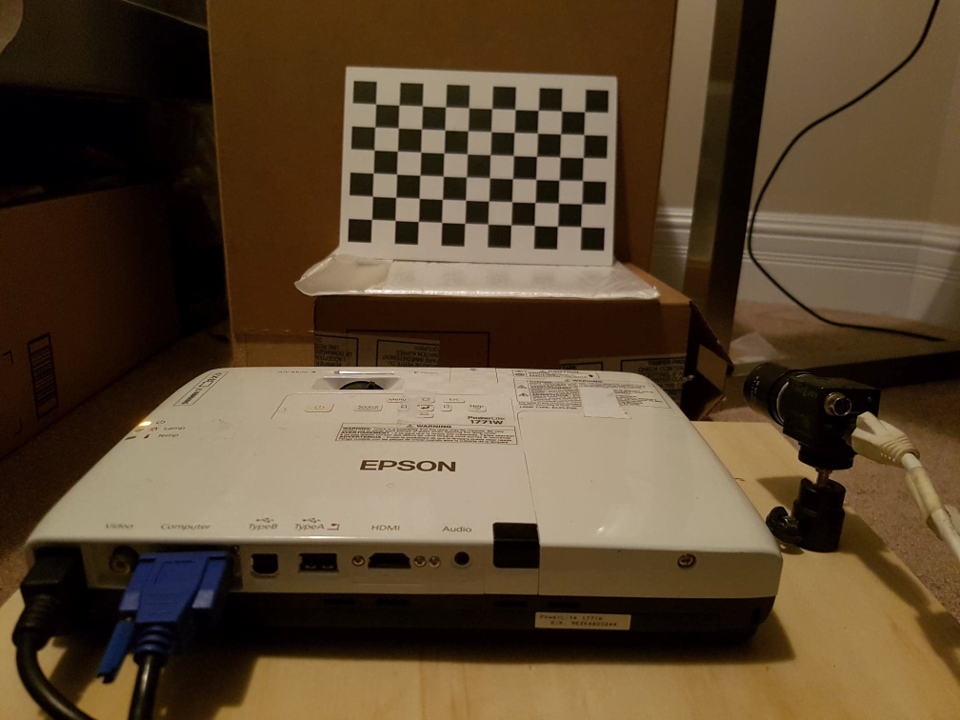}
	\caption{Procam system setup.  The projector (white) and the camera (black) are placed in front of the chessboard target. }
	\label{fig:procamsystem}
\end{figure}

This work proposes a user friendly method for calibrating a procam system from a single pose of a planar chessboard pattern of known dimensions as shown in Figure~\ref{fig:procamsystem}.  Camera-chessboard and projector-chessboard homographies are established with camera and chessboard after the camera and projector correspondences have been identified.  Using their respective homographic relationship with the chessboard both the camera and projector project their principal points onto it thereby establishing a definition for their respective principal axis.  Once their principal axis are defined their respective projection parameters are optimized using the Levenberg-Marquardt algorithm.  Once the projection relationship the chessboard has with the camera and projector is calculated it is trivial to compute the extrinsic parameters that relate the camera location to the projector location.  The user does not need to move any object or require any special purpose or expensive equipment.  

The accuracy of the recovered parameters from this method are comparable to 
those derived from mainstream techniques, all of which require multiple repositioning of a calibration target. To our best knowledge, this is the first example of procam calibration from a single pose of a planar target, which is the main contribution of this paper.

A second contribution is the implementation of a PnP-based technique for determining the precision of a group of calibration parameters.  This was included to enhance the reprojection error metric, which may not necessarily indicate how stable the final parameters are when used to determine arbitrary 3D points in space. The benefit of this technique is revealed from  the experimental results, which characterize the accuracy of the method and  demonstrates that it compares favorably with other more standard approaches. 



%% file: Background.tex
\section{\uppercase{Background}}
\label{sec:background}
There are many types of procam calibration methods,  all of which require one or more of the following;
\begin{itemize}
	\item 
	Images of a 2D target in several poses;
	\item A pre-calibrated camera;
	\item
	A precise electro-mechanical actuator, or;
	\item
	A 3D object that satisfies specific shape and detail constraints.
\end{itemize}
\vspace{-2pt}
Each of these requirements increases complexity and therefore decreases accessibility for the user, especially outside of a lab environment, as well as increasing potential sources of error. 

Methods based on Zhang's flexible calibration approach \cite{paper:zhangsMethod} are the most common, due to its effectiveness and popularity for camera calibration.  The main difference between such methods are the structured light technique used to acquire projector correspondences, and the patterns used on the 2D planar target, which tend to be chessboard corners~\cite{paper:CenterLine}\cite{paper:SingleShot}, circles~\cite{paper:PhaseError}\cite{paper:CircleEdge} or sometimes QR codes~\cite{paper:Fiducials} and random dot patterns~\cite{paper:DotsRand}.  Whichever structured light technique or 2D planar target pattern  is used, this type of calibration requires at least three poses of the 2D target plane to be imaged by the procam system~\cite{paper:zhangsMethod}.  It is an exacting and time consuming task to orient a planar target in multiple unique positions, while ensuring that it remains prominently in the fields of view of both the camera and the projector.

A method called \emph{visual servoing} can be used to calibrate a projector given a pre-calibrated camera.  The projector is to project a chessboard onto a physical one such that the corners of the physical and projected chessboard align~\cite{paper:VisSerBIG}.  This is done my modelling the projector as a virtual camera whose pose can be altered and is viewing the chessboard though the actual projector remains in the same position throughout the calibration process.  Using control theory the virtual camera is moved so that it is in the same pose as the projector that it is modelling, once the virtual camera and projector occupy the same position the projected chessboard will align with the physical one~\cite{paper:VisSerBASICS} \cite{paper:VisSerSMALL}.  Despite the camera (which is effectively half of the procam system) being precalibrated, at least 10 distinct poses of a chessboard are needed to calibrate the projector intrinsic and the extrinsic parameters.  This therefore has the same drawbacks as Zhang's method applied to procam calibration.

It is possible to calibrate a procam system if the position of a planar target can be precisely controlled.  This allows  Tsai's camera calibration method \cite{paper:Tsais} to be repurposed for procam calibration. In Tai's method,  calibration can be achieved with only two poses of a planar target, under the condition that these poses are related by a pure translation, and that the accurate translation value is known~\cite{paper:SameOrigin} \cite{paper:zhangsMethod}.  This can only be done if one has access to a programmable actuator, which severely limits the accessibility of this method.

Through the decomposition of a radial fundamental matrix and utilizing Bougnoux's equations, it is possible to simultaneously calibrate the projector and camera \cite{paper:SimSelfCal}.  Methods based on the aforementioned matrix and equations only require a 3D (i.e. non-planar) object imaged in a single pose to complete the calibration process \cite{paper:SimFund}.  However, for the values to be known to a global scale, the dimensions of the object must be known, and because it is based on a fundamental matrix, the object has shape and detail requirements that must be met for the results to be stable,
including that it should not be rotational or translation invariant and provide enough geometric variation to facilitate successful optimization \cite{paper:SimGeo}.

%% file: Method.tex
\section{\uppercase{Method}}
\label{sec:method}

The proposed method makes use of an error metric based upon mapping a set of known 2D coordinates on a planar chessboard pattern onto each of the camera and projector planes.  This linear mapping comprises the respective homographies between the two pairs of planes (i.e. chessboard to camera, and chessboard to projector) as well as the projection matrices formed by the camera and projector intrinsic and extrinsic parameters.  In the case of the camera,  a non-linear transformation of the image is also applied, to correct for radial distortions. 

The error metric is used to drive a non-linear optimization process over a set of parameters that model the projections.  The optimization is done separately for each of the camera and the projector, and its successful convergence  is reasonably robust to the selection of initial values.  Upon convergence, the resulting optimized parameters convert directly to establish the intrinsic and extrinsic parameters of the camera and projector, the accuracy of which is then independently evaluated with a separate test.
\subsection{Model and Notation}
\newcommand\numberthis{\addtocounter{equation}{1}\tag{\theequation}}

\begin{figure*}
	\centering
	\includegraphics[width=0.85\textwidth]{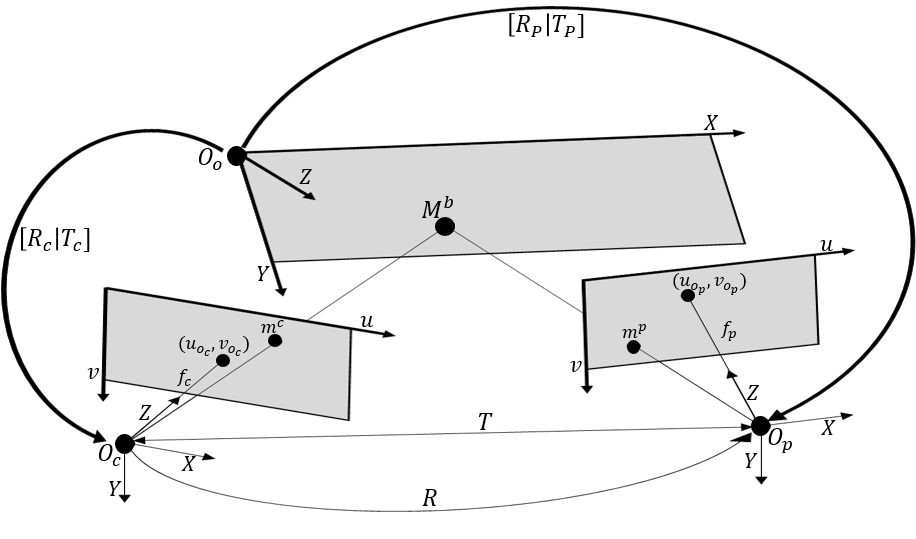}
	\caption{Procam System Model.}
	\label{fig:stereo}
\end{figure*}

The pinhole model is used to describe the camera and projector \cite{book:MultiViewGeo} as shown in Figure~\ref{fig:stereo} , with respective intrinsic matrices $K_c$ and $K_p$ defined as:
\begin{align*}
K_c &= \begin{bmatrix}
f_c&0&{u_o}_c\\
0&\alpha_c f_c&{v_o}_c\\
0&0&1\\
\end{bmatrix}, &
K_p &= \begin{bmatrix}
f_p&0&{u_o}_p\\
0&\alpha_p f_p&{v_o}_p\\
0&0&1\\
\end{bmatrix}  \numberthis
\end{align*}
Here, $f_c$ and $f_p$ represent the camera and projector focal lengths , and $[{u_o}_c  \ \ {v_o}_c]^T$ and $[{u_o}_p  \ \ {v_o}_p]^T$ are their respective principal points. The aspect ratio between the camera's $u$- and $v$-axes is denoted $\alpha_c$. The projector's aspect ratio is assumed to be unity, due to the high uniformity of commercial projectors, and so $\alpha_p = 1$ and this term is excluded from further consideration.
Camera lenses can exhibit significant radial distortion~\cite{paper:ParamFreeDist}, which can be estimated and corrected using the two parameter division model~\cite{paper:KevDistortion} that was introduced by Fitzgibbon~\cite{paper:DistModel}.  Alternately, projector lenses tend to be high quality and considered to possess relatively negligible amounts of radial distortion~\cite{paper:GeoPro}, and so we do not include  projector radial distortion in our calibration model.

Let $m^c = [u^c  \ \ v^c]^T$ and $\hat{m}^c = [\hat{u}^c  \ \ \hat{v}^c]^T$ represent corresponding undistorted and distorted 2D points in the camera image plane. The division model
transforms $\hat{m}^c$ to $m^c$
through the following:
\begin{gather*}
\begin{cases}
u^c = {u_o}_c + \dfrac{\hat{u}^c-{u_o}_c}{1+k_1r^2+k_2r^4}\\[8pt] \numberthis
v^c = {v_o}_c + \dfrac{\hat{v}^c-{v_o}_c}{1+k_1r^2+k_2r^4} 
\end{cases}
\label{eq:radist}
\end{gather*}
where $k_1$ and $k_2$ are the free parameters that need to be estimated, and $r$ is the Euclidean distance between the principal point and the distorted point.

The intrinsic matrices of the camera and projector allow 3D points defined in their respective coordinate systems to be transformed to their respective image planes as follows: 
\begin{align*}
sm^c &= K_cM^c &
sm^p &= K_pM^p \numberthis
\label{eq:sm=KM}
\end{align*}
where $m^j = [u^j  \ \ v^j]^T$ are 2D points on the camera ($j=c$) or projector ($j=p$) plane, and $M^j= [x^j  \ \ y^j  \ \ z^j]^T$ are 3D points defined in their corresponding coordinate reference systems as shown in Figure~\ref{fig:stereo}.  Points on the planar chessboard calibration target are defined in the chessboard coordinate frame, but are not \emph{a priori} defined in either the camera or projector coordinate frames.  To map these points from the chessboard frame to the camera or projector frames, the rotation and translation values between the chessboard and pinhole devices need to be determined.  Let $[R_c|T_c]$ and $[R_p|T_p]$ represent the rotation and translation of the chessboard in the camera and projector frames respectively, with the rotation matrices parameterized by the XYZ-Euler angles $(\psi_c, \nu_c, \phi_c)$ and $(\psi_p, \nu_p, \phi_p)$.  Multiplying the rotation and translation matrices by the corresponding intrinsic matrix allows for points defined in the chessboard coordinate system to be projected onto the camera and projector image planes: 
\begin{align*}
sm^c &= K_c[R_c|T_c]M^b &
sm^p &= K_p[R_p|T_p]M^b \numberthis
\end{align*}
where $M^b= [x^b  \ \ y^b  \ \ z^b]^T$  is a 3D point on the chessboard plane described in the chessboard frame as shown in Figure~\ref{fig:stereo}. 

The chessboard by construction lies on the $x^b$-$y^b$ plane, and so $M^b= [x^b  \ \ y^b  \ \ 0]^T$. Let $m^b=[x^b  \ \ y^b]^T$ represent a 2D coordinate on the chessboard plane such that $M^b= [{m^b}^T  \ \ 0]^T$.  The metric dimension of the chessboard is known, and its $N$ corner coordinate values are stored as $\{m^b_i\}_{i=1}^N$. The value of any $m^b_i$ is then related to the corresponding 2D $m^c_i$ and $m^p_i$ points through a pair of 2D homographies: 
 
\begin{equation}
m^b_i = H_C^Bm^c_i = H_P^Bm^p_i
\label{eq:H}
\end{equation}

Two sets of six parameters, 
denoted by $\boldsymbol{\theta_c}$
for the camera and 
$\boldsymbol{\theta_p}$
for the
projector, need to be determined to fully calibrate the procam system: 
\vspace{0.0 cm}
\begin{gather*}
\begin{cases}
\boldsymbol{\theta_c}&= \{f_c, \alpha_c, \phi_c, x_o^c, y_o^c, z_o^c \}\\[8pt] \numberthis
\boldsymbol{\theta_p}&= \{f_p, {v_o}_p, \phi_p, x_o^p, y_o^p, z_o^p \}
\end{cases}
\end{gather*}
Here, $O_c\!=\![x_o^c  \ \ y_o^c  \ \ z_o^c]^T$, $\phi_c$,\ and $O_p\!=\![x_o^p  \ \ y_o^p  \ \ z_o^p]^T$, $\phi_p$,\ denote the center of projection and the rotation about the principal axis of the camera and projector respectively.

\subsection{Camera Calibration}
The radial distortion present in the camera lens must first be calculated before $\boldsymbol{\theta_c}$ can be optimized.  This is accomplished by first extracting the epipole from the fundamental matrix between $\hat{m}^c_i$ and $m^b_i$, which is known as the \emph{fundamental matrix of radial distortion} for which the epipole is equivalent to the center of distortion \cite{paper:ParamFreeDist}.  The center of distortion extracted from this technique is used as the principal point for the camera.  Next, the distortion coefficients $k_1$ and $k_2$ are computed using the one-shot method of \cite{paper:KevDistortion}, which is based on the assumption that the homography between the $b_i$ and $m^c_i$ is proportional to the mapping between $\hat{m}^c_i$ and $m^c_i$.  Once $k_1$ and $k_2$ are calculated the undistorted camera image points $m^c_i$ can be recovered using Eq.~\ref{eq:radist}. 

\begin{figure}
	\centering
	\includegraphics[width=0.45\textwidth]{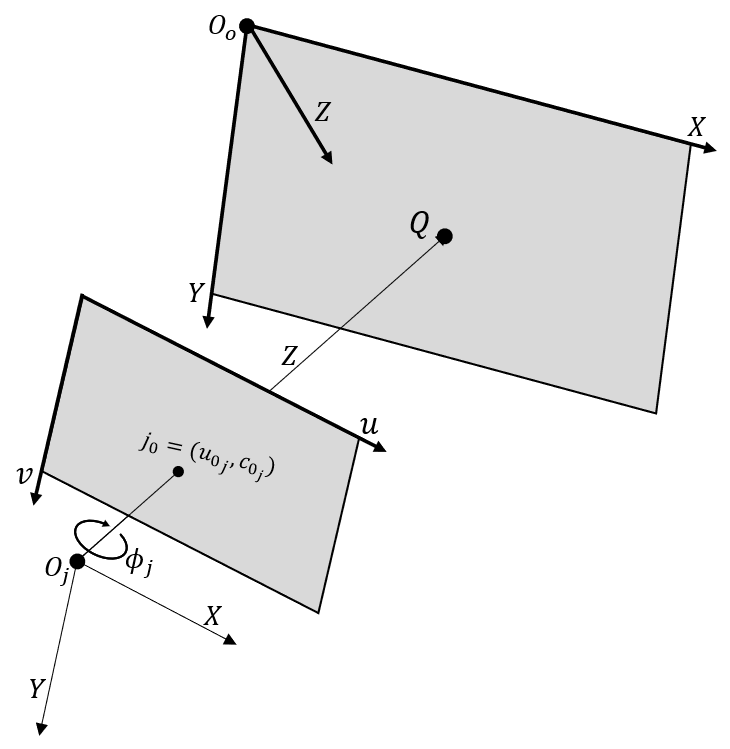}
	\caption{Formation of Principal Axis. For the camera, $j=c$ and $Q=C_o^B$. For the projector,  $j=p$ and $Q=P_o^B$.}
	\label{fig:ppaxis}
\end{figure}

Having calculated and corrected the camera radial distortion, the first step in optimizing $\boldsymbol{\theta_c}$ is to define the principal axis of the camera.  First the homography $H_C^B$ between the camera and chessboard is calculated
from Eq.~\ref{eq:H} using the known $m^b_i$ and the extracted and undistorted  $m^c_i$. Homography $H_C^B$ is then used to project the camera frame principal point $[{u_o}_c  \ \ {v_o}_c]^T$ onto the chessboard plane.  Let $C_o^B$ be the 3D location of $[{u_o}_c  \ \ {v_o}_c]^T$ projected onto the chessboard, expressed in the chessboard coordinate frame.  By definition the principal axis is the z-axis $\vec{Z}$, which originates at the origin of the camera reference frame, and intersects with the image plane at point $[{u_o}_c  \ \ {v_o}_c]^T$ and at the chessboard at point  $C^B_o$ as shown in Figure~\ref{fig:ppaxis}.
A rotation matrix $A$  is formed by the following:
\begin{gather*}
\begin{cases}
\vec{Z} &= C^B_o - O_c \\
\vec{Y} &= \vec{Z}\times[1  \ \ 0  \ \ 0]^T \\\numberthis
\vec{X} &= \vec{Y}\times\vec{Z}
\end{cases}
\end{gather*}

\begin{equation}
A = \begin{bmatrix}
\dfrac{\vec{X}}{|\vec{X}|}&\dfrac{\vec{Y}}{|\vec{Y}|}&\dfrac{\vec{Z}}{|\vec{Z}|}\\
\end{bmatrix}
\end{equation}
The directions of the $x$- and $y$-axes are controlled by $\phi_c$, which rotates them about the $z$-axis, creating a new rotation matrix $A_c=AR_Z(\phi_c)$. Next, $A_c$ and
$O_c$ 
are then used to calculate the extrinsic values between the chessboard and the camera:
\begin{align*}
R_c &= A_c^T, &
T_c &= -A^T\times O_c\numberthis
\label{eq:camcal}
\end{align*}

The location of $C^B_o$ is constant throughout the entire optimization process, therefore the camera is always pointed towards the same location on the chessboard plane.  Thus, as the value of $x_o^c$ changes the the camera is rotated about its $y$ axis and as the value of $y_o^c$ changes the camera is rotated about its $x$ axis.  There is no need to include $\psi_c$ and $\nu_c$, the rotations about the camera's $x$ and $y$ axis respectively, in $\theta_c$. 

The corners of the chessboard are then projected onto the camera image plane using Eq.~\ref{eq:sm=KM} with each $m^c_i$ replaced by $m_i(\boldsymbol{\theta_c})$.  The absolute difference between the undistorted camera image points $m^c_i$ and the transformed image points $m_i(\boldsymbol{\theta_c})$ is the error metric used to optimize $\boldsymbol{\theta_c}$, by applying the Levenberg-Marquardt algorithm to minimize the following cost function with respect to $\theta_c$:
\begin{equation}
\theta_c^* = 
\mbox{argmin}_{\theta_c}
\sum_{i=1}^{N}||m^c_i-m_i(\boldsymbol{\theta_c})||^2
\label{eq:camLM}
\end{equation}

\subsection{Projector Calibration}
As with the camera, the principal axis of the projector must be defined to optimize $\boldsymbol{\theta_p}$.  Homography $H_C^B$ is then used to project $[{u_o}_p  \ \ {v_o}_p]^T$ onto the chessboard plane, so that  $P_o^B$ is the 3D point of intersection of the projector optical axis and the chessboard plane, expressed in the chessboard coordinate system  as shown in Figure~\ref{fig:ppaxis}.  ${u_o}_p$ will always be equal to half the width of the projector image plane.  As before with the camera, and analogous to Eq.~\ref{eq:camcal}, a rotation matrix $A$ matrix is formed using Eq.~\ref{eq:projcal}, albeit with $\phi_c$ replaced by $\phi_p$.

Here, the directions of the $x$- and $y$-axes are controlled by $\phi_p$, which rotates them about the $z$-axis, creating a new rotation matrix $A_p=AR_Z(\psi)$. The $A_p$ and $O_p$ values are then used to calculate the extrinsic parameters between the chessboard and camera:
\begin{align*}
R_p &= A_p^T, &
T_p &= -A^T \times O_p \numberthis
\label{eq:projcal}
\end{align*}

The location of $P^B_o$ is confined to a line segment on the chessboard plane throughout the entire optimization process.  Thus, as the value of $x_o^p$ changes the the projector is rotated about its $y$ axis and as the values of $v_{op}$ and $y_o^p$ change the projector is rotated about its $x$ axis.  There is no need to include $\psi_p$ and $\nu_p$, the rotations about the projector's $x$ and $y$ axis respectively, in $\theta_p$.

The corners of the chessboard are then projected onto the projector image plane using Eq.~\ref{eq:sm=KM}, with $m_i^p$ replaced by $m_i(\boldsymbol{\theta_p})$.  The absolute difference between the projector image points $m_i^p$ and transformed image points $m_i(\boldsymbol{\theta_p})$ is the error metric used to optimize $\boldsymbol{\theta_p}$.  Analogous to the application of Eq.~\ref{eq:camLM} for the camera parameters, the Leveberg-Marquardt algorithm is then applied to minimize the following cost function with respect to the projector parameter set $\theta_p$:
\begin{equation}
\theta_p^* = 
\mbox{argmin}_{\theta_p}
\sum_{i=1}^{N}||m^p_i-m_i(\boldsymbol{\theta_p})||^2
\label{eq:projLM}
\end{equation}

%

\subsection{Projector Camera Calibration}
Lastly the procam extrinsic parameters are recovered from $[R_c|T_c]$ and $[R_p|T_p]$:
\begin{equation}
\begin{split}
R &= R_p R_c^T \\
T &= T_p-R\;T_c 
\label{eq:procamRT}
\end{split}
\end{equation}

\subsection{Initial Values}
\label{sec:Initial Values}
Initial estimates of the values of the two calibration parameter sets are required to commence the optimization process, with the values used listed in 
Table~\ref{tab:InitVals}.
The initial values of the focal lengths are the diagonal pixel length of their respective image planes \cite{paper:SimFund} where $r_c$, $r_p$, $c_c$, $c_p$ are the number of rows and columns in the projector and camera image planes.  The value of $\alpha_c$ is initialized to 1 as experimentally the aspect ratio is typically close to unity for pinhole devices \cite{paper:GreyLocalHom}.  The vertical coordinate  of the projector principal point, ${v_o}_p $, is usually near the top or bottom of the image plane, and so it was initialized to half the height of the projector image plane.  The values of $O_c = O_p = [0 \ \ 0 \ \ 2w_b]^T$ and $\phi_c =  \phi_p = 0^{\circ}$ were initialized based on the reasonable assumption that the chessboard is facing both the camera and the projector, and that their $x$- and $y$-axes are similarly oriented.  The width of the chessboard is represented by $w_b$.

\begin{table}
	\begin{center}
		\caption{Calibration Parameter Set Initial Values.}
		\begin{tabular}{c c} 
			\multicolumn{2}{c}{Camera}\\
			\hline
			$\boldsymbol{\theta_c}$ & Initial Value \\ [0.5ex] 
			\hline\hline
			$f_c$ & $\sqrt{r_c^2+c_c^2}$ \\ 
			\hline
			$\alpha_c$ & 1 \\
			\hline
			$\psi_c$ & 0  \\
			\hline
			$x_o^c$ & 0 \\
			\hline
			$y_o^c$ & 0  \\
			\hline
			$z_o^c$ & $2w_b$  \\ [1ex] 
			\hline
		\end{tabular}
		\quad
		\begin{tabular}{c c} 
			\multicolumn{2}{c}{Projector}\\
			\hline
			$\boldsymbol{\theta_p}$ & Initial Value \\ [0.5ex] 
			\hline\hline
			$f_p$ & $\sqrt{r_p^2+c_p^2}$ \\ 
			\hline
			$v_p$ & 
			$r_p/2$ \\
			\hline
			$\psi_p$ & 0  \\
			\hline
			$x_o^p$ & 0 \\
			\hline
			$y_o^p$ & 0  \\
			\hline
			$z_o^p$ & $2w_b$  \\ [1ex] 
			\hline
		\end{tabular}
		\label{tab:InitVals}
	\end{center}
\end{table}

\subsection{Correspondence Acquisition}
\label{sec:corralgo}
\begin{figure}
	\centering
	\includegraphics[width=0.23\textwidth]{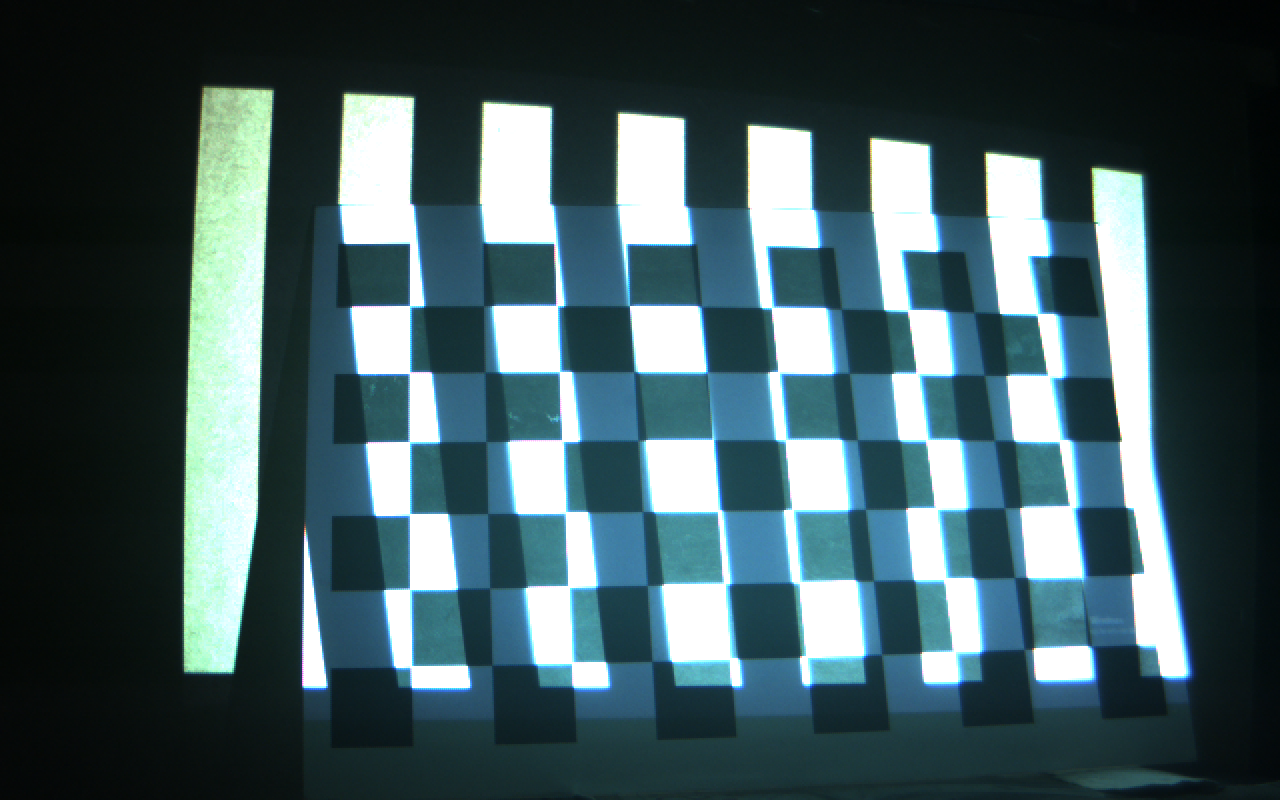}
	\includegraphics[width=0.23\textwidth]{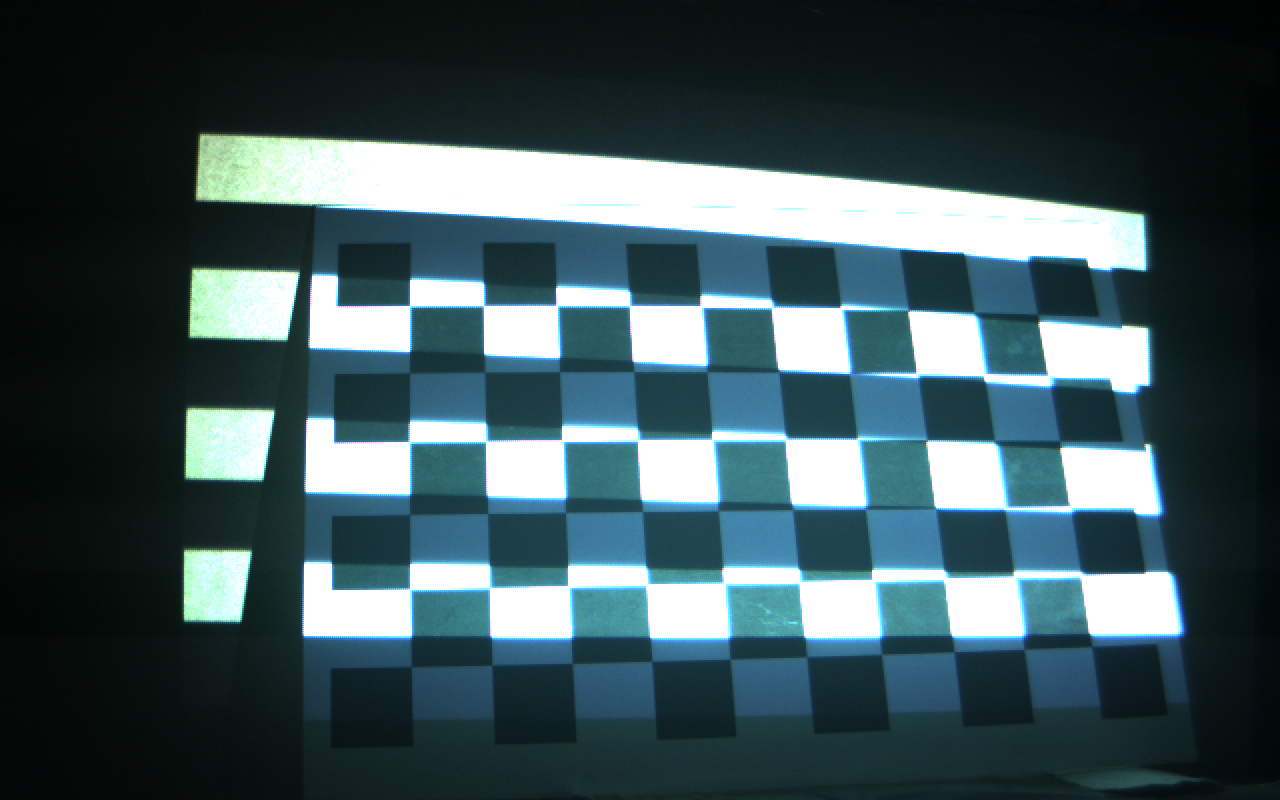}
	\caption{Horizontal and vertical Gray Codes projected onto chessboard.}
	\label{fig:gc}
\end{figure}

\begin{figure}
	\centering
	\includegraphics[width=0.23\textwidth]{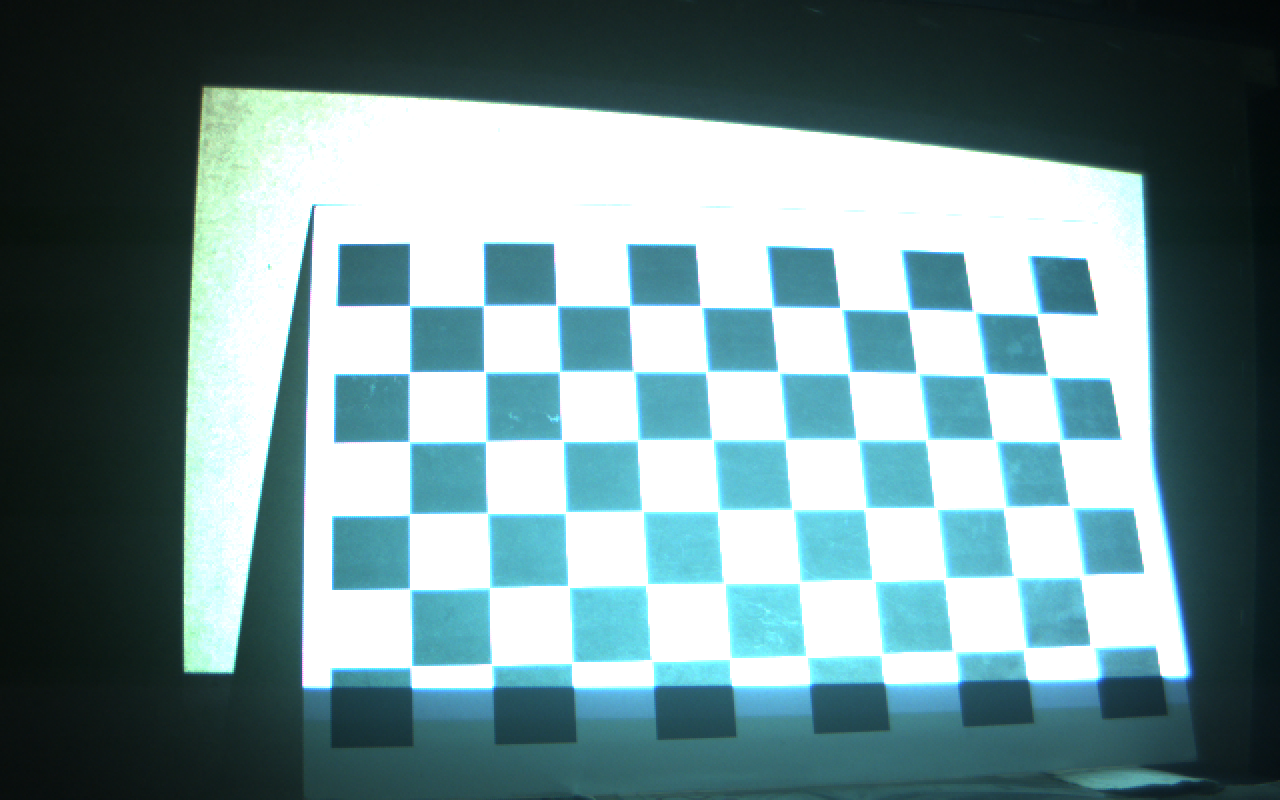}
	\includegraphics[width=0.23\textwidth]{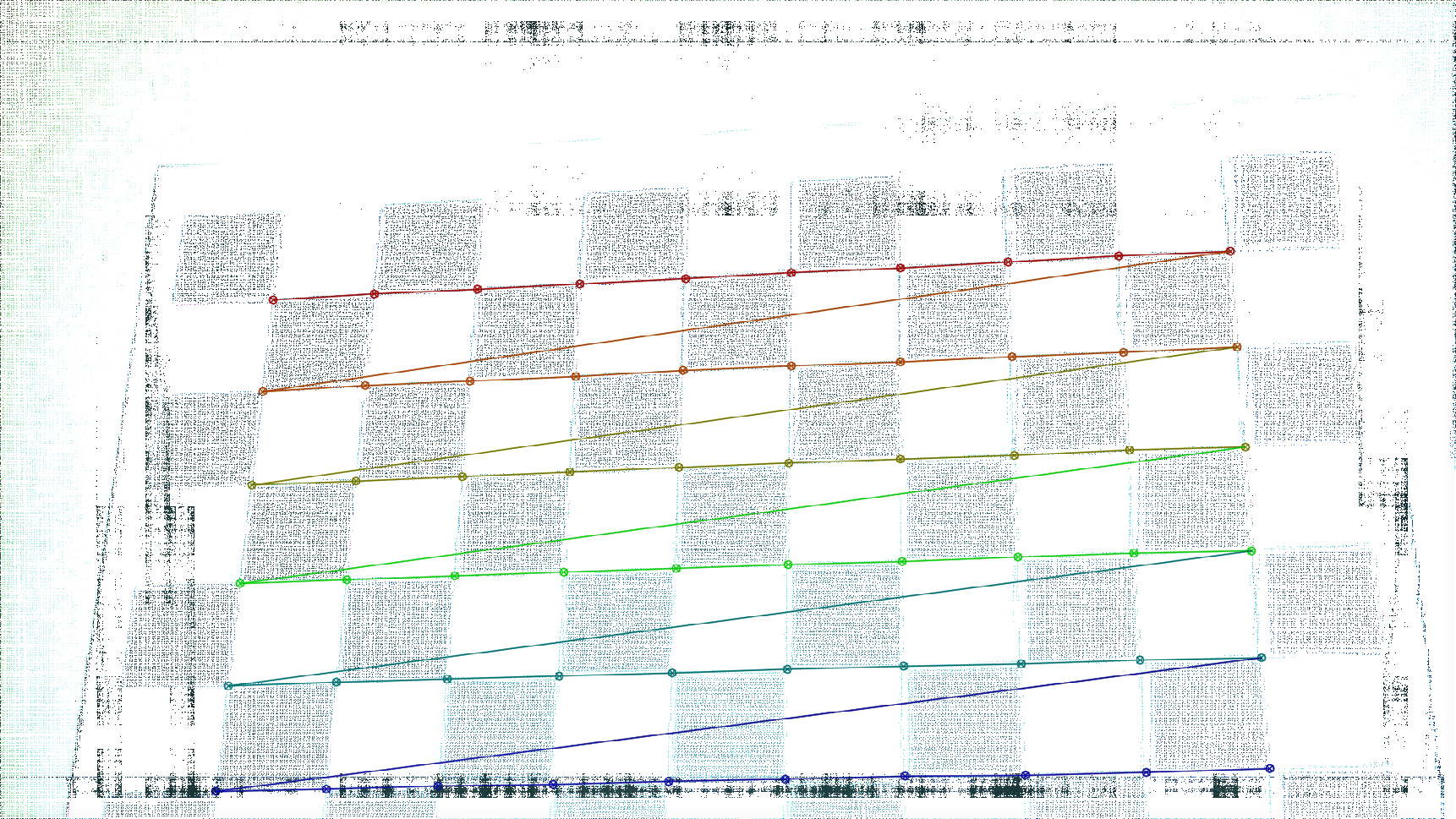}
	\caption{Camera Image (left) and projector image (right).}
	\label{fig:camproim}
\end{figure}

The $N$ chessboard corner coordinates are extracted to subpixel accuracy from the camera image and are stored in the array $\{\hat{m}^c_i\}_{i=1}^N$, with the $\;\hat{}\;$ symbol indicating that these coordinates exhibit the radial distortion, i.e. they have not yet been undistorted by applying Eq.~\ref{eq:radist}.  The projector frame coordinates corresponding to each of these camera frame coordinates are acquired by decoding a sequence of 46 projected Gray Code patterns, two of which are shown in Figure~\ref{fig:gc} and utilizing local homographies in the same manner as \cite{paper:GreyLocalHom}. The projector correspondences are then stored in the array $\{m^p_i\}_{i=1}^N$, ordered such that each $\hat{m}^c_i$, $m^p_i$ and $m^b_i$ correspond.  Figure~\ref{fig:camproim} shows the raw camera image and the computed projector `image'.

%% file: Experimentation.tex
\section{\uppercase{Experiments}}
\label{sec:exp}
Two sets of experiments were performed to test the accuracy of the proposed procam calibration method. One utilized a real procam system programmed with OpenCV in C++, and the other was a simulation programmed in MATLAB.  The real experiments show that different poses affect the accuracy of the calibration, and the simulated experiments allow us to isolate and identify what elements of the chessboard pose are the most influential factors that impact accuracy.  
\subsection{ProCam Setup}
The proposed calibration method was applied to a physical procam system. The camera used for this experiment was a BlackFly Point Grey color camera with a resolution of $1280\!\times\!800$ and the projector was the Epson PowerLite 1771W with a resolution of $1920\!\times\!1080$ as shown in Figure~\ref{fig:procamsystem}.  A chessboard of dimensions $21\!\times\!26.5$ cm with $10\!\times\!6 = 60$ corners was used as the calibration plane.  The chessboard was moved to seven distinct poses where the Gray Code sequence was projected, to establish sets of correspondences between the camera and projector points (described in Sec.~\ref{sec:corralgo}).

\subsection{Accuracy Metrics and Reference Values}
To assess the accuracy and validity of the calibration parameters, the calibration method of~\cite{paper:GreyLocalHom} was used to establish reference values to compare against.  They effectively used Zhang's method~\cite{paper:zhangsMethod} to calibrate both the camera and the projector.  Zhang's method requires minimally three distinct poses of a planar surface to calibrate a pinhole device, with 46 Gray Code images acquired for each chessboard pose to automatically establish correspondences between projector and camera coordinates.

As seven distinct poses of the chessboard were acquired, there are multiple groups of calibration parameters that can arise, each resulting from a unique combination of 
the seven distinct chessboard poses. 
Each such combination is referred to here 
as a \emph{pose set}. Each pose set
contains between three and seven chessboard poses, and so the number of pose sets is:
\begin{equation}
\sum_{i=3}^{7}\dfrac{7!}{i!(7-i)}= 99
\label{eq:combo}
\end{equation}

The reprojection error was used to determine the validity and of the calculated calibration parameters, as well as the two quantities $\sigma_{|T|}$ and $\sigma_{T}$, where $\sigma_{T}$ is the standard deviation of the projector location with respect to the camera, and $\sigma_{|T|}$ is the standard deviation of the projector's distance $|T|$ from the camera (known as the stereovision \emph{baseline}).  Assume the first group of calibration parameters were obtained using a combination of poses one, two and three. The intrinsic parameters from that pose set are used with the point correspondences extracted from the Gray Code images to calculate the $(X, Y,Z, |T|)$ values between the camera and projector, resulting in seven $(X, Y,Z, |T|)$ values, i.e. one for each pose, including the three poses (one, two and three) used to calculate the parameters. The $\sigma_{T}$ value is equal to the square root of the sum of the variances of the $(X, Y,Z)$ coordinates, and $\sigma_{|T|}$ is equal to the standard deviation of the $|T|$ values. Methods that solve the PnP problem can be used to extract the pose of pinhole device with respect to an object defined from a known set of planar or non-planar points if the intrinsic parameters are known~\cite{paper:EPnP} \cite{paper:GenPnP}.  Therefore OpenCV's solvePnP function is used in conjunction with Eq.~\ref{eq:procamRT} to calculate $\sigma_{|T|}$ and $\sigma_{T}$.

These standard deviation metrics are included in the analyses because reprojection error may not accurately predict the correctness of a possible 3D reconstruction.  No matter how the chessboard is moved and oriented in 3D space one should be able to calculate the same values for the procam extrinsic parameters using a PnP method because the pose between the camera and projector is rigid and thus constant in a procam system.  Thus when a PnP method is applied with the procam intrinsic parameters to calculate the extrinsics, $T$ and $|T|$ should remain constant, although a small amount of variation is expected due to noise.  Therefore the $\sigma_{|T|}$ and $\sigma_{T}$ measures the precision of the calibration.
\begin{table}
	\begin{center}
		\caption{Daniel and Gabriel Reprojection Errors.}
		\begin{tabular}{|p{1.3cm} |c| c| c|} 
			\hline
			\multicolumn{1}{|c|}{ }&
			\multicolumn{3}{c|}{Reprojection Error} \\
			\hline
			Stat & Cam & Pro & Stereo\\ [0.5ex] 
			\hline\hline
			Set A & 0.251 & 0.775 & 0.577 \\ 
			\hline
			Set B & 0.256 & 0.815 & 0.604\\ 
			\hline
			Mean & 0.274 & 0.855 & 0.637\\ 
			\hline
			Std Dev & 0.047 & 0.308 & 0.216\\  [1ex] 
			\hline
		\end{tabular}
		\label{tab:statsErrors}
	\end{center}
\end{table}

\begin{table}
	\begin{center}
		\caption{Standard Deviation of Baseline.}
		\begin{tabular}{|p{1.3cm} | c| c| c|c|c |} 
			\hline
			\multicolumn{1}{|c|}{ }&
			\multicolumn{5}{c|}{Translation Vector} \\
			\hline
			Stat & $\sigma_X$ & $\sigma_Y$ & $\sigma_Z$ & $\sigma_{T}$ & $\sigma_{|T|}$ \\ [0.5ex] 
			\hline\hline
			Set A &4.49 & 1.70 & 1.76 & 5.11& 2.97 \\ 
			\hline
			Set B & 0.79 & 0.90 & 0.53 & 1.31& 0.70 \\ 
			\hline
			Mean & 6.06 & 2.91 & 2.46 & 7.27& 3.86 \\ 
			\hline
			Std Dev & 5.53 & 3.11 & 2.07 & 6.55& 4.20\\  [1ex] 
			\hline
		\end{tabular}
		\label{tab:statsErrorsT}
	\end{center}
\end{table}

The reprojection errors, $\sigma_{|T|}$ and $\sigma_{T}$ were calculated for all 99 pose sets.  Pose set 99 was calibrated using all seven poses, and pose set 4 was calibrated using only poses four, five and six.  For convenience, these are referred to as Set A and Set B, respectively.  Set A produced the lowest reprojection error for the camera, projector and procam system, while the parameters of Set B resulted in the lowest values for $\sigma_{|T|}$ and $\sigma_{T}$ as shown in Tables~\ref{tab:statsErrors} and~\ref{tab:statsErrorsT}.  
The parameters of Set B are used as the reference values because translation standard deviation values are less than half of Set A's, and their resultant reprojection errors are within no more than 5\% of Set A's.  This indicates that the parameters resulting from Set B will be able to extract depth to a higher degree of accuracy than Set A for arbitrary objects because while Set A is marginally more accurate than Set B, Set A is significantly less precise than Set B.

\begin{table}
	\begin{center}
		\caption{Daniel and Gabriel Calibration Results.}
		\begin{tabular}{|p{1.3cm} |c| c| c|}
			\hline
			Parameter & Reference Values & Mean& Std Dev\\ [0.5ex] 
			\hline\hline
			$f_c$ & 1535.7 & 1541.8 & 57.4\\ 
			\hline
			$\alpha_c f_c$ & 1537.9 & 1542.3 & 54.7\\ 
			\hline
			${u_0}_c$ & 666.4 & 666.4 & 10.1\\ 
			\hline
			${v_0}_c$ & 518.3 & 518.3 & 21.6\\ 
			\hline
			$f_p$ & 2506.4 & 2428.2 & 179.4\\ 
			\hline
			$\alpha_p f_p$ & 2507.2 & 2430.7 & 157.7\\ 
			\hline
			${u_0}_p$ & 1007.9 & 996.1 & 20.4\\ 
			\hline
			${v_0}_p$ & 1046.0 & 1074.7 & 50.1\\ 
			\hline
			$\psi$ & 0.87 & 1.82 & 1.42\\ 
			\hline
			$\nu$ & 14.44 & 14.91 & 0.71\\ 
			\hline
			$\phi$ & -0.25 & -0.12 & 0.34\\ 
			\hline
			X & -170.05 & -174.12 & 6.57\\ 
			\hline
			Y & -41.25 & -38.37 & 6.49\\ 
			\hline
			Z & -65.35 & -77.90 & 36.32\\ 
			\hline
		\end{tabular}
		\label{tab:statsKc}
	\end{center}
\end{table}

\subsection{Real Data Results}
Each of the seven chessboard poses produced a different final group of procam calibration parameters when the proposed method was used.  The values of the parameters is largely dependent on the orientation between the pinhole devices and the chessboard.  Table~\ref{tab:CalKc} and \ref{tab:CalKp} show the calculated intrinsic parameters of the camera and projector respectively for each pose. 
\begin{table}
	\begin{center}
		\caption{Camera Intrinsic Parameters.}
		\begin{tabular}{| c c c c c|} 
			\hline
			Set & $f_c$ & $\alpha_c f_c$ & ${u_0}_c$ & ${v_0}_c$ \\ [0.5ex] 
			\hline\hline
			1 & 1492.91	& 1497.45 & 653.77 & 542.00 \\ 
			\hline
			2 & 1400.22	& 1402.12 & 624.69 & 549.09\\ 
			\hline
			3 & 1907.85	& 1911.66 & 646.03 & 543.90 \\ 
			\hline
			4 & 1585.58	& 1589.60 & 676.94 & 517.70\\ 
			\hline
			5 & 1484.22 & 1492.50	& 677.57 & 516.98\\ 
			\hline
			6 & 1493.04 & 1494.36 & 621.42 & 526.89 \\ 
			\hline
			7 & 2139.43 & 2142.98 & 611.36 & 560.52\\ 
			\hline
		\end{tabular}
		\label{tab:CalKc}
	\end{center}
\end{table}
\begin{table}
	\begin{center}
		\caption{Projector Intrinsic Parameters.}
		\begin{tabular}{|c c c c  c|} 
			\hline
			Set & $f_p$ & $f_p$ & ${u_0}_p$ & ${v_0}_p$ \\
			\hline\hline
			1 & 2812.21	& 2812.21 & 960.00 & 908.01 \\ 
			\hline
			2 & 3364.91	& 3364.91 & 960.00 & 820.48\\ 
			\hline
			3 & 2537.86	& 2537.86 & 960.00 & 1065.43 \\ 
			\hline
			4 & 2586.15	& 2586.15 & 960.00 & 1042.14\\ 
			\hline
			5 & 2357.50 & 2357.50 & 960.00 & 1039.86\\ 
			\hline
			6 & 1893.04 & 1893.04 & 960.00 & 1159.24 \\ 
			\hline
			7 & 2554.36 & 2554.36 & 960.00 & 1077.02\\ 
			\hline
		\end{tabular}
		\label{tab:CalKp}
	\end{center}
\end{table}

The camera reprojection error is largely dependent on the sum of the absolute values of $\psi_c$ and $\nu_c$.  This pattern is clearly displayed in Table~\ref{tab:CamAccuracy} which also lists $|\Delta f_c|$, the absolute difference between the reference values and calculated focal length of the camera for each set.  Set one, five and six possess $|\psi_c|+|\nu_c|$ values over 25 degrees and their average reprojection error across the seven images is less then 0.4.  Sets 3 and 7 both have $|\psi_c|+|\nu_c|$ less than 10 degrees average reprojection error is greater than 1.  Set two and set four both were both calibrated with $|\psi_c|+|\nu_c|$ values that are greater than 10 degrees but less then 25 degrees, and their average reprojection errors are 0.42 and 0.44 respectively.  Generally the higher the $|\psi_c|+|\nu_c|$  value, the lower the reprojection errors that the intrinsic matrix will provide.  Also, the sets that provide three of  the lowest reprojection error, set one, four and six produce $|\Delta f_c|$ values that are less than the standard deviation value of $f_c$ in Table~\ref{tab:statsKc}.
\begin{table}
	\addtolength{\tabcolsep}{-2pt}
	\begin{center}
		\caption{Camera Calibration Accuracy Metrics.}
		\begin{tabular}{| c c c c c c|} 
			\hline
			Set & $\psi_c$ & $\nu_c$ &$|\psi_c|+ |\nu_c|$& $|\Delta f_c|$& Error  \\ [0.5ex] 
			\hline\hline
			1 & -19.11 & -8.32 & 27.43 & 42.78 & 0.30\\ 
			\hline
			2 & -12.25 & -9.42 & 21.68 & 135.47 & 0.42\\ 
			\hline
			3 & -4.47 & 4.92 & 9.40 & 372.16 & 1.2\\ 
			\hline
			4 & -9.62 & 1.74 & 11.36 & 49.88 & 0.44\\ 
			\hline
			5 & -5.69 & -29.44 & 35.12 & 51.47 & 0.39\\ 
			\hline
			6 & -14.18 & -15.64 & 29.82 & 42.65 & 0.31 \\ 
			\hline
			7 & -4.13 & 5.01 & 9.14 & 603.73 & 1.62\\
			\hline
		\end{tabular}
		\label{tab:CamAccuracy}
	\end{center}
\end{table}

The projector reprojection error is also dependent of the chessboard rotation about its x- and y-axis, but in a different manner than the camera.  Sets three, four, five and seven possess $|\nu_p|$ that are all greater than 13 degrees, and each of their mean reprojection errors are less than 1.5.  Sets one, two and six all posses $|\nu_p|$ values less than 10 degrees and consequently their reprojection errors are all greater than 1.5 as displayed in Table~\ref{tab:ProAccuracy}.  Therefore generally the greater the $|\nu_p|$ value, the lower the associated reprojection errors that the intrinsic matrix will provide.  $|\Delta f_p|$, the absolute difference between the reference focal length and calculated focal length  is also displayed in Table~\ref{tab:ProAccuracy}, the sets that provide the lowest four reprojection errors (i.e. sets three, four, five and seven), produce $|\Delta f_p|$ values that are less than the standard deviation of $f_p$ in Table~\ref{tab:statsKc}.

\begin{table}
	\begin{center}
		\caption{Projector Calibration Accuracy Metrics.}
		\begin{tabular}{|c c c c c|} 
			\hline
			Set & $\psi_p$ & $\nu_p$& $|\Delta f_p|$ & Error  \\ [0.5ex] 
			\hline\hline
			1 & -22.03 & 7.05& 305.02 & 1.57\\ 
			\hline
			2 & -16.39 & 5.10& 857.72 & 2.81\\ 
			\hline
			3 & 1.69 & 14.46& 30.66 & 0.96\\ 
			\hline
			4 & -8.23 & 17.92 & 78.95 & 0.96\\ 
			\hline
			5 & -6.44 & -13.63 & 149.70 & 1.25\\ 
			\hline
			6 & -9.15 & -1.82 & 614.15 & 2.27\\ 
			\hline
			7 & -2.16 & 17.81 & 47.17 & 1.01\\ 
			\hline
		\end{tabular}
		\label{tab:ProAccuracy}
	\end{center}
\end{table}

\begin{table}
	\addtolength{\tabcolsep}{-3pt}
	\begin{center}
		\caption{ProCam Euler Angles and Translation Vector.}
		\begin{tabular}{|c c c c c c c|} 
			\hline
			\multicolumn{1}{|c}{ }&
			\multicolumn{3}{|c|}{Euler Angles}&
			\multicolumn{3}{|c|}{Translation Vector}\\
			\hline
			Set & $\psi$ & $\nu$ & $\phi$ & X & Y & Z \\ [0.5ex] 
			\hline\hline
			1 & -2.77 & 15.41 & -1.06 & -170.20 & -39.13 & 8.66 \\ 
			\hline
			2 & -5.13 & 14.22 & -1.76 & -163.19 & -39.14 & 124.57\\ 
			\hline
			3 & -1.69 & 14.46 & -0.53 & -201.81 & -23.20 & -186.90 \\ 
			\hline
			4 & 1.24 & 16.19 & -0.63 & -178.27 & -36.52 & -62.35 \\ 
			\hline
			5 & 1.66 & 15.74 & -0.25 & -164.90 & -34.38 & -73.10\\ 
			\hline
			6 & 3.63 & 14.23 & 0.86 & -170.68 & -35.38 & -155.81 \\ 
			\hline
			7 & 1.53 & 12.85 & 0.19 & -211.10 & -20.29 & -255.61 \\ [1ex] 
			\hline
		\end{tabular}
		\label{tab:CalRT}
	\end{center}
\end{table}

The accuracy of the extrinsic parameters which are displayed in Table~\ref{tab:CalRT} are dependent on the accuracy of the intrinsic parameters.  Intrinsic parameters dictate the location of the image plane relative to the center of distortion, and where the ray connecting a 3D point and the center of projection intercept the associated image plane.  For example, the Z value of set two and set seven diverge the most from the reference value, with values 189.92 and 190.26 mm respectively.  The projector focal length produced from set two deviates from the reference value the most compared to the six other projector focal lengths.  Likewise for the camera focal length produced by set seven.  Sets four and five produce two of the top four calibration parameters for the camera and the projector, and as a result have the most accurate Z value.

The results of the $\sigma_{|T|}$, $\sigma_{T}$ and the stereo reprojection error are highly correlated.  Generally, the closer the calibration parameters (both intrinsic and extrinsic) are to the reference value, the smaller the errors and translation standard deviations are, as shown in Table~\ref{tab:allTResults}.  Set one and three have the same stereo reprojection, but their $\sigma_{T}$ value differs by about 2 mm and their $\sigma_{|T|}$ value differ by about 7 mm.  This suggests that despite possessing the same correspondence error, the calibration parameters of set three would provide a better surface reconstruction than set one.

\begin{table}
	\begin{center}
		\caption{PNP Translation Vector and Stereo Reprojection Error Results.}
		\begin{tabular}{|c |c| c| c|} 
			\hline
			Set & Magnitude & Location& Stereo \\ [0.5ex] 
			\hline\hline
			1 & 10.45 & 11.40 & 1.13 \\ 
			\hline
			2 & 17.94 & 21.40 & 2.02 \\ 
			\hline
			3 & 3.68 & 9.64 & 1.13 \\ 
			\hline
			4 & 1.49 & 2.34 & 0.75 \\ 
			\hline
			5 & 1.48 & 2.99 & 0.93 \\ 
			\hline
			6 & 11.69 & 20.72 & 1.62 \\ 
			\hline
			7 & 6.74 & 18.62 & 1.40 \\
			\hline
		\end{tabular}
		\label{tab:allTResults}
	\end{center}
\end{table}

\subsection{Simulation Results}
From the results of the real procam system experiments, it is shown that that the pose of the chessboard relative to the camera and projector directly affect the calibration accuracy.  This simulated experiment isolates and examines this phenomenon more precisely, by rotating the projector and camera individually with respect to the chessboard to identify how the varying degrees of rotation affect calibration accuracy.  The value of $|\Delta f_c|$ and $|\Delta f_p|$ are highly correlated with their respective reprojection errors as shown in Table~\ref{tab:CamAccuracy} and \ref{tab:ProAccuracy} and are therefore used as the accuracy metrics for this experiment.  The simulated $K_c$ values are; $f_c=1539$, $\alpha_c= 1.004$, ${u_0}_c=674$, and ${v_0}_c=512$.  The simulated $K_p$ values are; $f_p=2421$, $\alpha_p= 1.002$, ${u_0}_p=1013$, and  ${u_0}_p=1065$.  The resolution of the simulated camera and projector are 
$1280\!\times\!800$
and 
$1900\!\times\!1080$
pixels respectively.  The simulated chessboard was $21.\times 26.5$ cm with $10\times 6= 60$ corners.  

Let $(\psi_c,\nu_c)$ and $(\psi_p,\nu_p)$ be the rotations about x- and y-axes for the camera and projector respectively.  The principal point to calibrate the camera was shifted 5 pixels to the right and downwards from its true value to account for the real world issue that the calculated center of distortion is not located in the exact same place as the ground truth principal point of the camera.  The assumptions about the projector intrinsic matrix in this section are consistent with the assumptions made in the previous section.  Throughout all rotations, the translation vector for both devices were kept constant, along with the rotations about the z-axis (i.e. $\phi$) as well as their respective intrinsic matrices.
\subsubsection{Camera}
\begin{figure}
	\centering
	\includegraphics[width=0.5\textwidth]{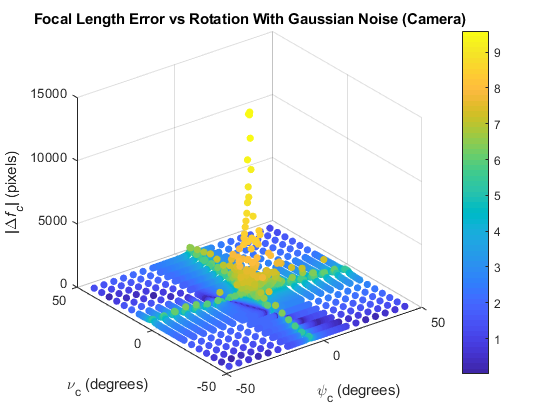}
	\caption{Camera Focal Length Error vs $\psi_c,\nu_c$}
	\label{fig:CamNoiseFlE}
\end{figure}

\begin{figure}
	\centering
	\includegraphics[width=0.45\textwidth]{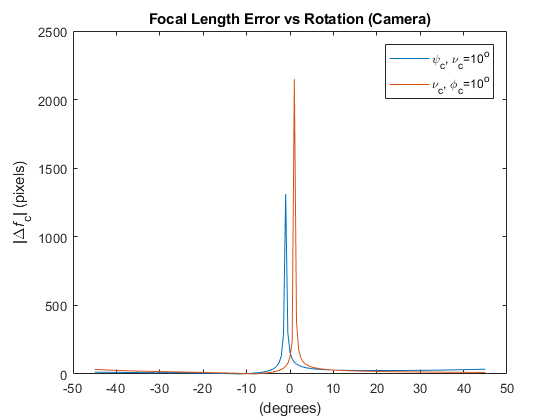}
	\caption{Rotate $\psi_c$, $\nu_c=10^{\circ}$ and $\nu_c$, $\psi_c=10^{\circ}$}
	\label{fig:cam2DRot}
\end{figure}

The value of $|\Delta f_c|$ generally decreases as $|\psi_c|+|\nu_c|$ increases, as shown in Figure~\ref{fig:CamNoiseFlE} where yellow points signify high error and dark blue points signify low error. Figure~\ref{fig:cam2DRot} shows a plot of $|\Delta f_c|$ vs. $\psi_c$, where $\nu_c=10^{\circ}\;\forall\;(\psi_c,|\Delta f_c|)$, and $|\Delta f_c|$ vs $\nu_c$ where $\psi_c=10^{\circ}\;\forall\; (\nu_c,|\Delta f_c|)$.  Despite the fact that for each of the aforementioned graphs one of the angles is fixed, they represent the overall relationship between the focal length error and  $(\nu_c, \psi_c)$, which is that $|\Delta f_c|$ exponentially decays as $|\psi_c|$ and $|\nu_c|$ increase.


\subsubsection{Projector}
\begin{figure}
	\centering
	\includegraphics[width=0.45\textwidth]{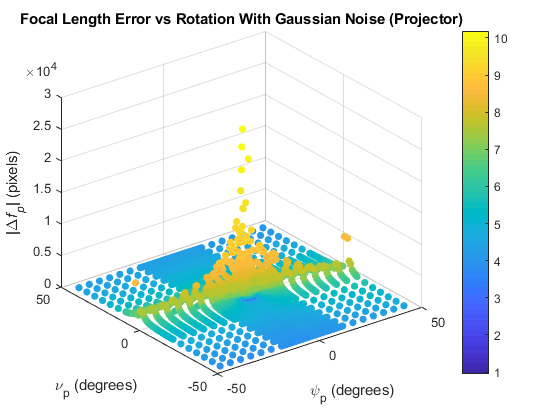}
	\caption{Projector Focal length Error vs $\psi_p,\nu_p$.}
	\label{fig:ProFlENoise}
\end{figure}

\begin{figure}
	\centering
	\includegraphics[width=0.45\textwidth]{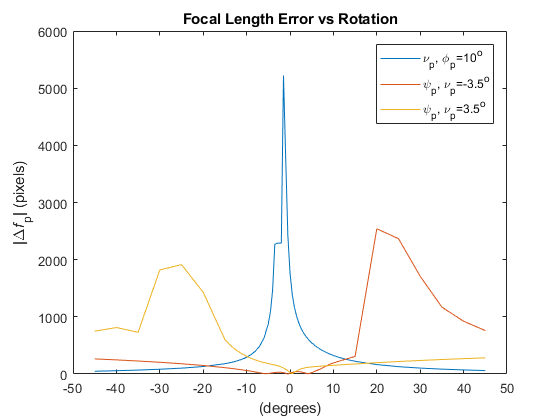}
	\caption{Rotate $\nu_p, \psi_p=10^{\circ}$; $\psi_p, \nu_p=-3.5^{\circ}$; and  $\psi_p, \nu_p=3.5^{\circ}$}
	\label{fig:Pro2DRot}
\end{figure}

The overall relationship between $|\Delta f_p|$ and ($\psi_p,\nu_p$) is not the same as the camera's due to the different set of assumptions regarding the intrinsic parameters. The  $|\Delta f_p|$ value still exponentially decays as $|\nu_p|$ increases for all $\psi_c=\theta$, as shown in Figures~\ref{fig:ProFlENoise} and~\ref{fig:Pro2DRot}.  However, the relationship between $|\Delta f_p|$ and $\nu_p$ changes depending on the corresponding $\psi_p$, as shown in Figures~\ref{fig:Pro2DRot}.  As $\nu_p$ is kept constant from -45 to 45 degrees, the relationship between  $|\Delta f_p|$ and $\psi_p$  transitions from generally increasing with $\psi_p$ , exponentially decaying with $|\psi_p |$ and generally decreasing with $\psi_p$.  

\subsection{Discussion}
The best intrinsic calibration results from the proposed method are comparable to the results of Daniel and Gabriel's technique.  The reprojection error of the camera using intrinsic values from set one and set six are within one standard deviation of the mean reprojection error from \cite{paper:GreyLocalHom}.  All sets where $|\psi_c|+|\nu_c| > 10^{\circ}$ resulted in reprojection errors below 0.45 for the camera.  The reprojection error of the projector using intrinsic values from set three, four and seven are within one standard deviation of the mean reprojection error from \cite{paper:GreyLocalHom} and all sets where $|\nu_p| > 13^{\circ}$ resulted in reprojection errors below 1.3 despite not accounting for distortion.  

The best procam calibration results from the proposed method are also comparable to the results of Daniel and Gabriel's technique.  All sets used to calibrate the system with the proposed technique except set two resulted in $\sigma_T$ values within a standard deviation of the mean value in Table~\ref{tab:allTResults}.  Set three, four, five and six, where at least either $|\psi_c|+|\nu_c| > 10^{\circ}$ or $|\nu_p| > 13^{\circ}$ achieved $\sigma_T$ values less than then the mean of the Zhang-style calibration method.  Sets four and five are the only ones that resulted in $\sigma_{|T|}$ values that are within one standard deviation of the value returned by Set B and are also less than the mean value.  They are also the only two sets where $|\psi_c|+|\nu_c| > 10^{\circ}$ and $|\nu_p| > 13^{\circ}$ and are therefore able to achieve a stereo reprojection error of less than 1. The stereo reprojection error from set four is also within one standard deviation of the reprojection error derived from the ground truth method in Table~\ref{tab:statsErrors}.

%% file: Conclusion.tex
\section{\uppercase{Conclusions}}
\label{sec:conclusion}
This work has proposed a simple and accurate method of calibrating a procam system.  It is 
very user friendly,
requiring only a single pose
of the 
planar chessboard target, without any requirement
to reposition the target to multiple poses.
It therefore
contains none of the practical drawbacks 
and inconveniences of mainstream techniques, while maintaining comparable reprojection errors and stability of the estimated parameters.  The conditions that provide the best results are easy to follow and are reproducible; orient the chessboard with respect to the camera and projector so that $|\psi_c|+|\nu_c|> 10^{\circ}$ degrees and $|\nu_p|> 13^{\circ}$ degrees, respectively.  Generally the calibration accuracy improves as both quantities increase.


